# Physics-Informed Machine Learning for Steel Development: A Computational Framework and CCT Diagram Modelling

Peter Hedström,[a,b,*] Victor Lamelas Cubero,[b] Jón Sigurdsson,[b] Viktor Österberg,[b] Satish Kolli,[c] Joakim Odqvist,[a,b] Ziyong Hou,[d] Wangzhong Mu,[e] and Viswanadh Gowtham Arigela,[b]

[a]KTH Royal Institute of Technology, Department of Materials Science and Engineering, Sweden.
[b]Ferritico AB, Sweden.
[c]Freudenberg Group, Germany.
[d]International Joint Laboratory for Light Alloys (MOE), Chongqing University, China.
[e]Engineering Materials, Luleå University of Technology, Sweden.
 *Correspondence should be addressed to P.H. (pheds@kth.se or peter.hedstrom@ferritico.com)

Machine learning (ML) has emerged as a powerful tool for accelerating the computational design and production of materials. In materials science, ML has primarily supported large-scale discovery of novel compounds using first-principles data, as well as digital twin applications aimed at optimizing specific manufacturing processes. However, developing and applying general-purpose ML frameworks to complex industrial materials, such as steel, remains a significant challenge. A major obstacle is accurately capturing the intricate relationship between chemical composition, processing parameters, and the resulting microstructure and properties. To address this gap, we introduce a computational framework that combines physical insights with ML to develop a physics-informed continuous cooling transformation (CCT) model for steels. Our model, developed using a dataset of unprecedented scale (4100 diagrams), is validated against both literature data and experimental results. The model demonstrates high computational efficiency, generating full CCT diagrams with 100 cooling curves in less than 5 s[#], and shows strong generalizability across alloy steels with high predictive accuracy, achieving phase formation classification F1 scores > 88% for all phases. For phase transition temperature regression, it attains mean absolute errors (MAE) < 20 °C across all phases except bainite, which exhibits slightly higher MAE of 27 °C. By extending the framework with additional generic and customized ML models, a platform can be established that serves as a universal digital twin for heat treatment and more. Integration with complementary simulation tools and targeted experiments will further support accelerated workflows.

## Introduction

Accelerating the discovery and design of advanced materials has become a central focus in materials science over the past decade, driven by the need for novel materials and processing solutions to address pressing global challenges. In addressing engineering challenges such as hydrogen infrastructure, vehicle electrification, and decarbonized manufacturing, steel remains a critical engineering material, with annual production nearly twice that of aluminum alloys, the second most widely used engineering metal. For complex industrial materials such as steel, Integrated Computational Materials Engineering (ICME) is now recognized as the state-of-the-art approach for industrial materials design [1]. It is widely accepted that an ICME framework emphasizing modeling and virtual testing can significantly reduce trial-and-error iterations during alloy design, thereby accelerating the development process for new steels [2].

Within the ICME framework, the ongoing development and refinement of physically based simulations using computational thermodynamics and kinetics [3], i.e. the CALPHAD methodology, have been instrumental in establishing ICME as a mainstream approach in industry. Other physically based modelling techniques that operate at smaller scales, such as density functional theory (DFT) and molecular dynamics (MD) calculations, have also been used to gain fundamental understanding of industrial materials [4]. However, such methods have limited applicability to engineering challenges where computational speed and the ability to account for a complex interplay of influencing parameters are critical. Pioneering work on statistical modelling using machine learning (ML) for steels was conducted by Bhadeshia and others in the 1990s [5-7], and recent advances in methodology, computational power, and data accessibility have led to a surge in ML applications addressing metallurgical research and development challenges [8]. ML approaches complement the physically based ICME modelling methods, providing a more suitable modeling approach in many cases, for example, when processes are highly complex and probabilistic, and a multivariate statistical framework is preferred that can naturally integrate uncertainties and stochastic variations.

While ML offers advantages in modeling complex systems, purely data-driven approaches often lack robustness in terms of generalizability and physical interpretability. To address this limitation, physics-informed ML integrates domain knowledge to enhance generalizability [9]. Physical knowledge can be incorporated in various ways, including: (1) Wang et al. [10] used CALPHAD-based predictions as input features in their ML modeling framework to predict the martensite start temperature (Ms) in steel; (2) Cao et al. [11] derived physically based features to include in their ML modeling framework of steel phase transformation products and their morphologies, and (3) Lee et al. [12] developed a framework where closed-form analytical expressions were used together with ML to predict the yield strength of metallic alloys.

Another strategy involves developing physics-informed ML models that explicitly encode physical constraints and known relationships among features (4). This approach not only embeds physics into the ML framework but also improves development efficiency by eliminating computational bottlenecks associated with traditional physically based models and enabling rapid retraining as new data becomes available. Importantly, the various physics-informed ML approaches are complementary and can be integrated into unified modeling frameworks.

In this manuscript, we present a physics-informed ML framework that is primarily aligned with the fourth strategy for integrating physical knowledge into ML, while remaining adaptable to other strategies. This framework is demonstrated through the development of a model for predicting continuous cooling transformation (CCT) behavior in steel. Although several CCT models have been proposed previously [13-19], they have typically relied on small databases, often limited to a few hundred alloys, and have not systematically incorporated physical principles. Our model is trained on data from 4100 alloys and embeds physical knowledge within the ML architecture to deliver a scalable and generalizable modeling framework. This foundation enables seamless integration with both generic and customized ML models, supporting the development of a universal digital platform for steel heat treatment. Following model validation against literature data and new experimental results, we discuss potential extensions and practical applications of the proposed physics-informed ML framework.

## Modelling methodology

In this section we will present the general data-driven modelling approach used to develop the CCT model. We describe the process of database construction, ML model development, and validation.

### Data Collection, and Pre-processing

For this work over 4100 CCT diagrams were ingested from the CCT and TTT Diagrams of Steels database [20] and its referenced sources. Since the HTML files that made up the database were generated using opaque JavaScript, the chemistry and heat treatment were scraped from the rendered files using the browser automation package Selenium [21] and a collection of Python routines. Graphical information from the diagrams was collected by tracking mouse-clicks on the figures. This resulted in a database containing over 37310 data-points, each corresponding to a measured cooling rate. In cases where the austenitization temperature was not explicitly provided, it was assumed to correspond to the intercept of the y-axis. The compositional landscape spans steel grades ranging from low-alloy steels to high-Cr stainless steels and refractory steels containing elevated levels of W (Table 1).

**Table 1:** Non-zero minimum (m) and maximum (M) values for all features. Units are as follows: °C for austenitization temperature (T), seconds for austenitization time (t) and logarithmic exponent of the cooling rate (CR) in °C/s. All alloying elements are expressed in wt.%.

|   | T | t | CR | C | Mn | Si | Cr | Mo | S | P |
|---|---|---|---|---|---|---|---|---|---|---|
| m | 600 | 0.1 | -3.03 | 3E-4 | 5E-6 | 7E-6 | 3E-6 | 8E-4 | 4E-5 | 7E-5 |
| M | 1480 | 9000 | 3.25 | 3.89 | 12.5 | 4.48 | 26 | 9.46 | 0.45 | 0.14 |

|   | Ni | Al | Cu | Nb | V | W | Co | N | Ti | B |
|---|---|---|---|---|---|---|---|---|---|---|
| m | 1E-5 | 0,001 | 1E-3 | 1E-4 | 3E-4 | 5E-3 | 2E-3 | 5E-4 | 3E-4 | 4E-6 |
| M | 20 | 2.1 | 6.5 | 3 | 10 | 17.8 | 13.4 | 1.8 | 1.62 | 0.03 |

Following data collection, data cleaning was performed to ensure dataset reliability. Initial cleaning steps included handling missing values, addressing outliers, and removing duplicates and faulty data samples. Missing values for the austenitization time were handled by imputing average values for that parameter. The composition, transformation temperatures, austenitization time and austenitization temperature, were scaled and shifted into distributions centered around 0 with standard deviation 1.

This processed data was then fitted to an autoencoder, a neural network with equal numbers of input and output nodes (see Figure 1), using an encoding/decoding architecture of 11-4-4-11 nodes and the tanh activation function in the bottleneck layers. By employing an hourglass-shaped neural network, with the inputs used as labels or targets, the model is forced to learn a low-dimensional representation of the data, exaggerating the difference between inconsistent data points and the main distribution. The autoencoder was trained using the Adam optimization algorithm as implemented in TensorFlow [22] until the training error had converged to within 3 digits. By analyzing the reconstruction error, most diagrams that had been erroneously digitized in the purchased database [20] could be identified and corrected. However, some inconsistent datapoints for cooling rates above 100 °C/s remained and the 1490 data points with the highest reconstruction error were considered unreliable and removed from the dataset. After data cleaning, the final dataset contained 35820 data-points. From this dataset, 10% of the data points were randomly selected and sequestered as a test set. To ensure the independence between training and test sets, the split was performed at the diagram level, such that all data points originating from the same CCT diagram were grouped together, preventing any single diagram from contributing to both subsets.

The remaining data was scaled from its original form using quantile-based scaling to facilitate the learning process for certain algorithms applied in this study (i.e.: KNN and SVM). Each sample includes information about composition, austenitization temperature, cooling rate, phase formation temperatures, phase fractions and phase presence.

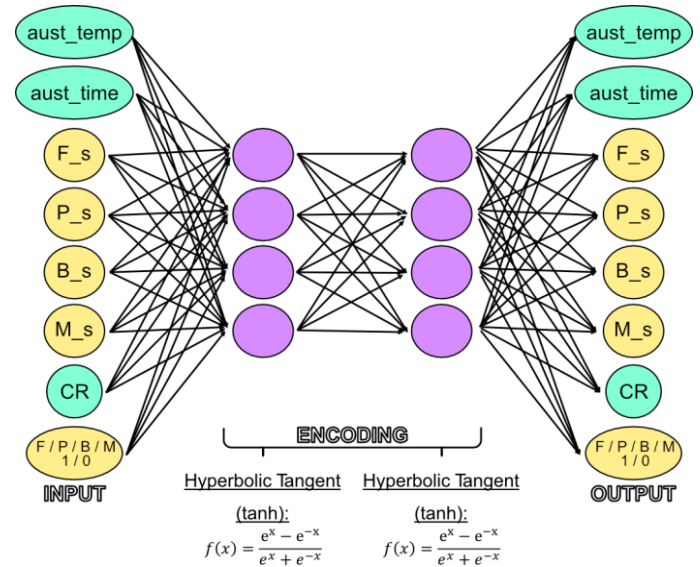

**Figure 1:** Schematic illustration of the autoencoder used for outlier detection.

**Machine learning approach**

The CCT module is conceptualized as a combination of models, where each sub-model predicts a distinct component of the CCT diagram. The problem is decomposed into a set of classification and regression tasks. Classification models predict the presence of the microconstituents namely Ferrite, Pearlite, Bainite, and Martensite. Regression models aim at finding the transformation temperatures for a given sample. A stacked modeling strategy is used, where the output of one model affects the other models' output. In this case, the regression tasks' results depend on the classification task's output. Hence, the CCT model comprises 14 sub-models (see Figure 2), communicating with each other when relevant and generating necessary elements to display a complete CCT diagram. The chosen strategy to integrate the different models requires specific steel domain knowledge and shows how the accuracy of ML tools can benefit from including physics-informed guidance in their predictions.

taken as the average of the errors for the three models against each validation set. The hyperparameters chosen to create the final model, using all the training data, were those that minimized the validation error. Subsequently, the models were evaluated on the test set to provide an unbiased estimate of their respective performance.

The machine learning models evaluated in this framework are widely established and therefore only briefly described here. The k-Nearest Neighbours (KNN) algorithm predicts values based on the average outcome of the closest data points in the feature space [25]. Support Vector Machines (SVM) identify an optimal hyperplane that separates classes with maximal margin to enhance classification accuracy [26].

The rest of the models (Random Forests, Gradient Boosting & Extreme Gradient Boosting) [24, 27-28] are all made up of a set of decision trees but differ in how they are structured and their intended purpose. Random Forests are made up of independent, randomized trees whose predictions are aggregated. The boosting frameworks, however, work in a sequential manner, where each base learner

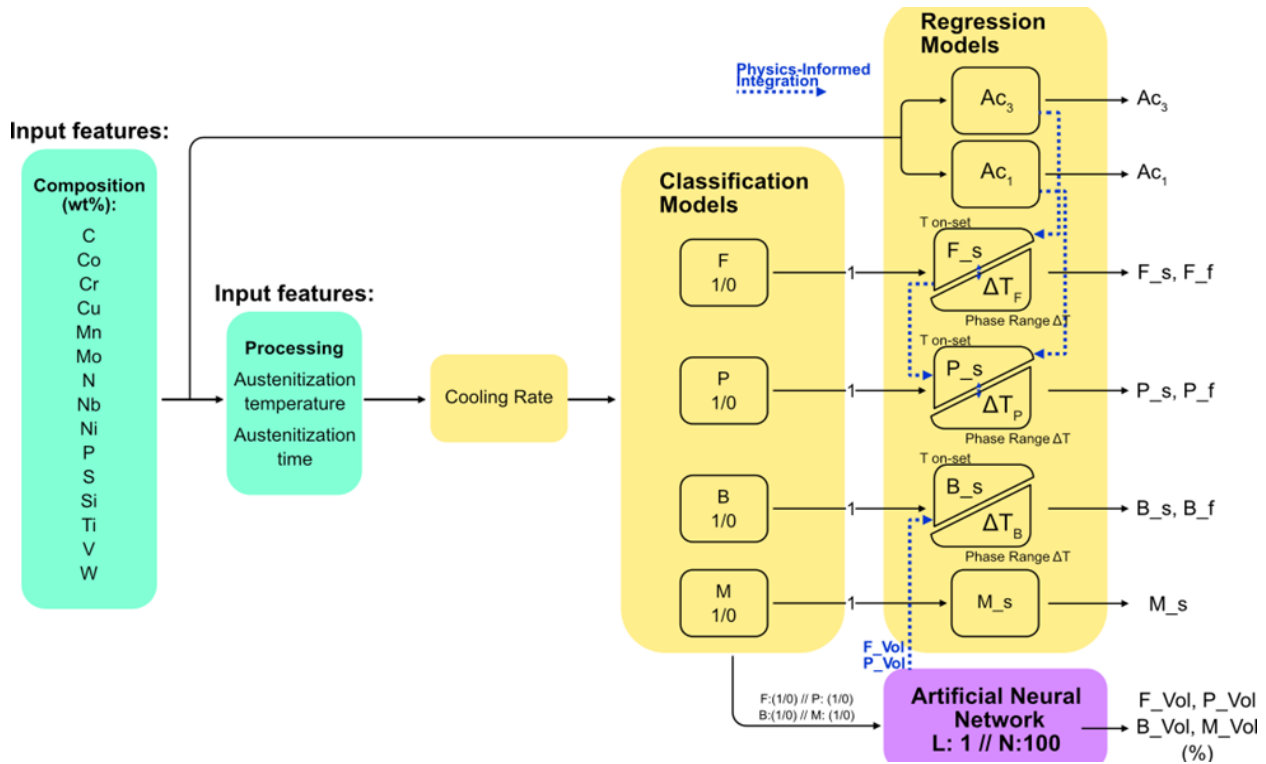

**Figure 2:** CCT model architecture is set up by 14 sub-models.

*Phase existence and Transformation temperatures*

The k-Nearest Neighbors (K-NN), Support Vector Machines (SVM), Random Forests (RF) and Gradient Boosting (GB) models were implemented using Scikit-Learn [23]. Extreme Gradient Boosting (XGB) models were also generated using the official Python module [24]. Each of these learning algorithms has a different set of hyperparameters, which need to be properly configured to get the best results out of the model. For this purpose, cross-validation was used. Cross-validation involves splitting the training data into subsets and letting each subset serve as a validation set for a model trained on the remaining data. In this work, the training data was split into three folds. When evaluating hyperparameters, three models were created according to the splits, and the validation error was attempts to correct the prediction of the previous learner.

These models and the above-mentioned strategy were used to train classifiers for each phase and regressors for the transformation temperatures, including $Ac_1$ and $Ac_3$. Rather than modeling the temperature at which the transformation ends, the duration of the transformation was modeled.

*Phase fractions*

The modelling of phase fractions followed a multiclass classification strategy. A neural network with one hidden layer containing 100 nodes was employed. The hidden layer used the relu6 = min (max (0, x), 6) activation function. The output layer used no bias to ensure that phases which were not formed would be encoded as = 0 % already in the hidden layer. The output layer used the SoftMax activation function F(x) where x denotes final layer outputs prior to activation (equation 4). This guarantees that the output nodes are smooth functions and that their sum adds up to 100%. The neural network model is illustrated in Figure 3.

$$F(x) = \frac{e^x}{\sum_{j=1}^{K} e^{xj}} \quad (4)$$

10% of the training data was isolated as a validation set, separate from both the training and test set. The model was trained using the Adam optimization algorithm as implemented in TensorFlow [22] until the validation error showed no improvement for three consecutive epochs.

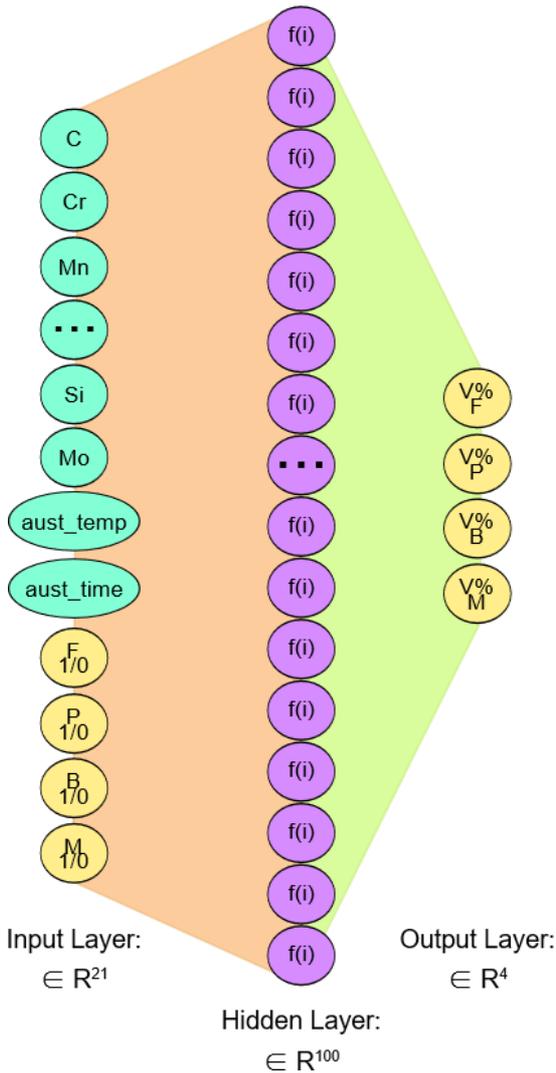

**Figure 3:** Schematic showing the neural network used for phase fraction prediction of ferrite, pearlite, bainite and martensite.

*Physics-informed integration*

The introduction of physical metallurgy principles into the modelling framework aims to enhance the predictive power of the models. Physics can be incorporated into the modeling process in various ways. One approach involves the engineering of features based on domain-specific knowledge, and in this study, the definition of an austenitization parameter that combines data on austenitization time and temperatures was tested. Both features influence austenitization kinetics in the same direction, as well as having the same kind of microstructural effect on the prior austenite grain size (PAGS). Thus, the simplification of the feature space by combining two features into one while keeping the feature values more continuous and diverse across the whole search space was intended to improve model performance. However, no significant improvement was observed when using the combined factor compared to treating temperature and time as separate features; therefore, they were retained independently.

Another feature engineering option to enhance the physical guidance during model training is to include direct information on the PAGS. The approach implemented in [19] to model PAGS and similar variations of it, was tested without significant success. The challenge remains the estimation of the grain size prior to the soaking step (rarely reported); such data would allow for the definition of an initial state to be used in physics-based grain growth models and determine PAGS for all austenitization conditions. Such an implementation remains a promising topic for future work in feature engineering.

Beyond feature engineering, several physics-informed integration strategies based on metallurgical knowledge were introduced to improve model accuracy and ensure physical consistency:

1. Synthetic data generation for phase formation (classification models). To address the discrete nature of measured CCT diagrams, synthetic data on phase formation was introduced between two consecutive experimental cooling rates that both exhibit the presence of a given phase. Since only a limited number of cooling curves are typically reported in experimental diagrams, this interpolation expands the dataset available for training classification models. Based on metallurgical principles, it is reasonable to assume that phase formation occurs continuously between two cooling rates where the phase is observed, allowing the model to better capture transformation boundaries and improve classification robustness.

2. Synthetic data points at cooling rates approaching zero were introduced for the ferrite start ($F_s$) and pearlite start ($P_s$) models. These data were generated using the outputs of the $A_C3$ and $A_C1$ models, respectively, as under near-equilibrium conditions, $A_C3$ corresponds to the ferrite start temperature and $A_C1$ to the pearlite start temperature, assuming both were measured at very slow heating rates. This integration enforces thermodynamic coherence between equilibrium transformation temperatures and those predicted under finite cooling rates.

3. For all diffusional transformations (Ferrite, Pearlite and Bainite), the transformation start temperature was included as an input feature for predicting the transformation range. From a metallurgical perspective, a higher transformation start temperature implies increased atomic mobility and faster diffusion, which can accelerate the transformation and result in a narrower temperature range. Including this dependency helps the model capture fundamental kinetic relationships that are typically observed experimentally in CCT.
4. The output of the ferrite start model ($F_s$) was coupled with the pearlite start model ($P_s$) to reflect the sequential nature of diffusional transformations in steels. Since ferrite generally forms before pearlite during cooling and both transformations are compositionally and kinetically linked, this coupling constrains the model to follow physically realistic transformation sequences. It also prevents unphysical outcomes, such as pearlite forming prior to ferrite, which could otherwise arise from purely data-driven predictions.
5. The presence of ferrite and pearlite prior to the bainitic transformation has an impact on the composition of the remaining austenite that later transforms into bainite during cooling. This influences the bainitic transformation temperature as well as the extent of the transformation. To account for this effect, the NN-predicted ferrite and pearlite fractions ($F_{frac}$ and $P_{frac}$) were included as additional input features for the bainite start temperature ($B_s$) model. For instance, if one takes the extreme case where the bainite classifier predicts bainite formation, but the phase fraction NN predicts $F_{frac}+P_{frac} = 1$, the former prediction would be overridden and the calculation for $B_s$ would not be used.

The bainite finish temperature ($B_f$) was set equal to the martensite start temperature ($M_s$) whenever the predicted $B_f$ was lower than $M_s$. Although isothermal bainite formation has been reported below $M_s$, under continuous cooling conditions such transformations are highly unlikely, as the onset of the martensitic, diffusionless transformation rapidly consumes the remaining austenite.

## Results and Discussion

### Classification of phases

Figure 4 displays the F1-scores for the predictions of the different models on phase existence when evaluated on the test set. The F1-score is employed as the evaluation metric because it balances precision and recall, offering a more informative measure than accuracy alone. As can be seen from Figure 4, the Gradient Boosting (GB) classifiers seem to outperform the other models on the test set.

With the results from Figure 4 in mind, the Gradient Boosting (GB) models were chosen for further analysis. The confusion matrices for the classifiers predicting the existence of each phase are shown in Figure 5. The classification models generally have very good accuracy, precision, and recall, which means that the models perform well in predicting when a phase forms and when it does not. On average, an accuracy of 88% is seen for all classification models.

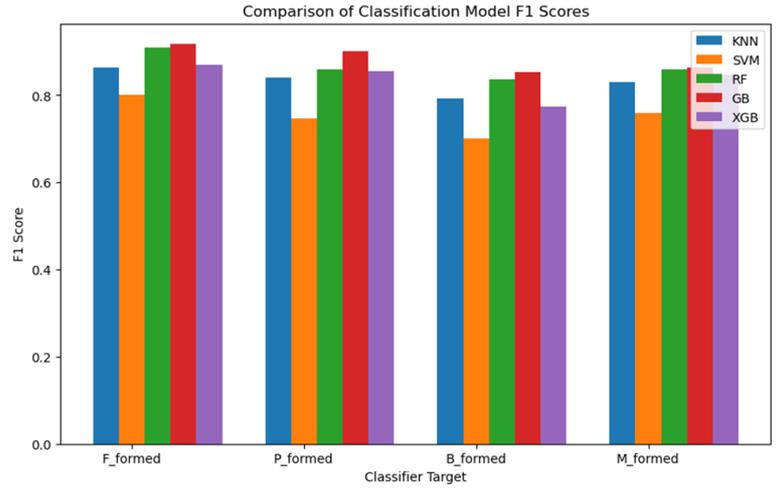

**Figure 4:** F1-scores for the different trained models when predicting phase formation in the test set.

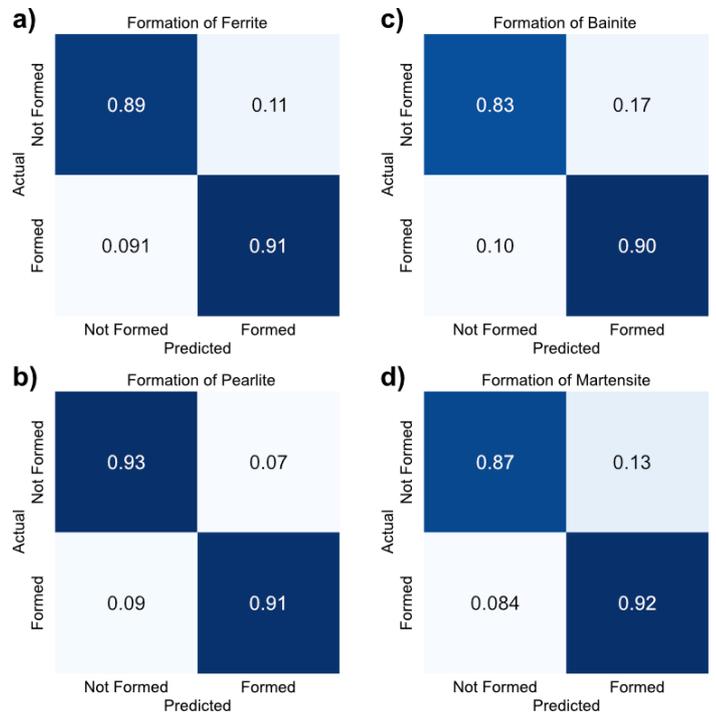

**Figure 5:** Confusion matrices with row normalization for classification models of a) ferrite b) pearlite c) bainite d) martensite formations.

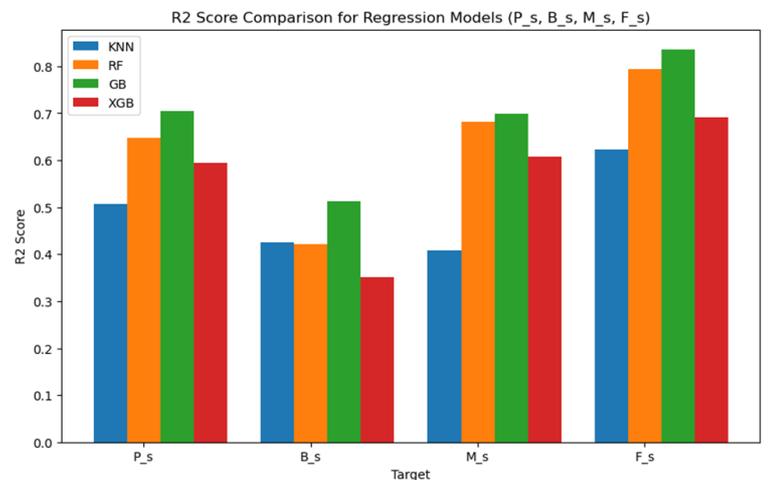

**Figure 6:** R2 scores for the different models on phase start temperature predictions.

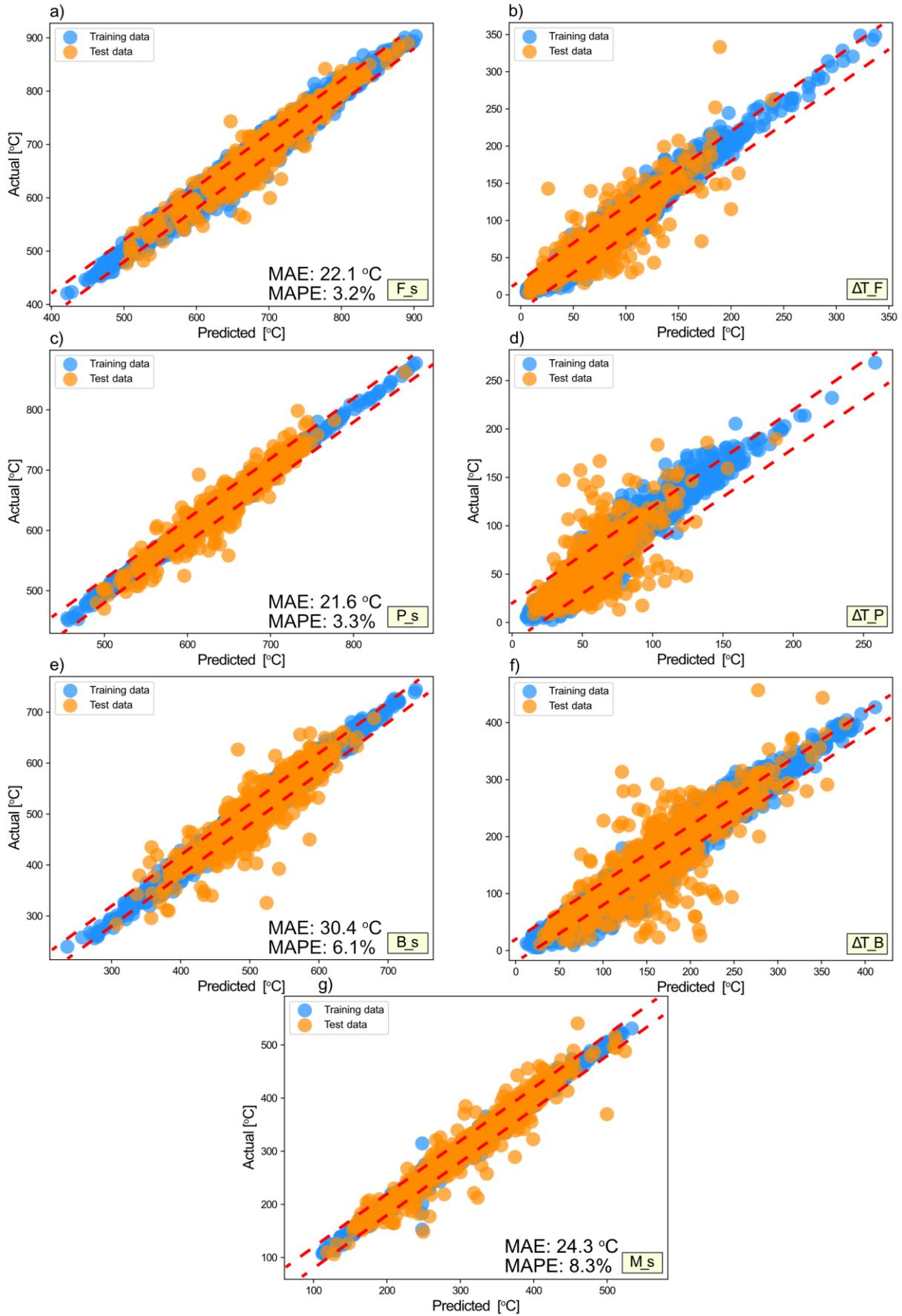

**Figure 7:** Measured vs predicted phase transformation start temperatures for a) ferrite, c) pearlite, e) bainite and g) martensite. And Measured vs predicted phase transformation progression temperature for b) ferrite, d) pearlite and f) bainite.

**Regression analysis of transformation temperatures and phase fractions**

Figure 6 depicts the $R^2$-scores for the predictive performance of the different models on the phase start temperatures in the test set. As is highlighted in the figure, the Gradient Boosting (GB) models seem to outperform the others on this dataset. The same is true for the other regression models. The GB models were therefore selected as the best candidates for further analysis and predictions.

For phase formation temperatures, the model seems to perform very well for ferrite, pearlite, and martensite with an average error around 20°C. However, there is still room for improvement in bainite predictions as the error rises to 30 °C. Similar studies using smaller datasets [19] have shown potential improvement in the accuracy of bainite transformation temperatures predictions by increasing, even further, the coupling between regression and phase fraction models at the model's training and integration step. The calculated onset and offset temperatures of ferrite are displayed against the experimental values in Figure 7a. The dispersed plot's linear form implies that predicted and measured values are in high agreement. Similar behavior is observed for the phase fractions when evaluating the proposed model on the test data. For ferrite, and pearlite (Figure 7c), an average percentage error of less than 5% is noted. However, for bainite and martensite (Figure 7e and g), the mean absolute percentage error is closer to 10%, possibly resulting from current integration not fully capturing the previous thermal history of the material when bainitic or martensitic transformations start. When comparing the R-square values (Figure 6), a similar result is observed. The models can explain 80% and 73% of the variance for ferrite and pearlite, respectively. On the other hand, the model explains 55% of the variance in the data for bainite.

In Figure 7, the predicted transformation temperatures are plotted against the measured temperatures. The validation against an unseen test dataset could already serve as a reasonable benchmarking of the models. However, to gain even further insight into the accuracy of the final model collection, 34 additional CCT diagrams not included in the main used database [20], and completely foreign to the training and testing data set, were gathered from 11 articles [17, 29-38] and used to further benchmark the model. The compositions and the corresponding mean absolute errors for all the alloys used in the benchmarking can be found in Table S.1. The errors are generally similar to those seen in the previously evaluated test data predictions, and similar accuracy is attained for alloys with high (i.e. ID 27) and low (i.e. ID 10) alloying content. Similar to what is observed in the test set validation, the microstructure with a lower accuracy in the benchmark predictions is bainite. The detailed analysis for the diagrams showing the highest bainite start errors ($B\_s^{Err}>50°C$) suggests the possible sources for such deviations as well as relevant next steps in the modeling of CCT diagrams. On one side, ID 26 in Table S.1 presents the highest MAE for Bs. Such sample in reference [37] is deformed prior to cooling, making the thermomechanical history of the sample more complex than the standard conditions used to train the models, thus this presents a source of uncertainty. An expanded CCT modeling would account for hot forming processes prior to cooling in the prediction of phase transformation. On the other side, multiple samples from reference [31] also stand out for having $B\_s^{Err}>50°C$, these samples have in common the systematic iteration of B content. B is an element that is well known to affect the bainite kinetics very strongly, in the present modelling framework the variation of ppm's of B in the composition seems to be not fully captured. To aim at this inaccuracy, additional feature engineering on the B content modeling to capture more accurately its effect, i.e. including cross-feature interaction parameters that include B, could be considered. With the same objective, one could also consider developing specialized models using the same methodology as the current one but trained on a B specific dataset to increase model accuracy. The implementation of such domain-specific models highlights the higher flexibility of ML and data-driven methodologies with respect to physically based simulations, being the former able to adapt to new material systems without the need of extensive database recalibration once the modeling pipeline and the data are available.

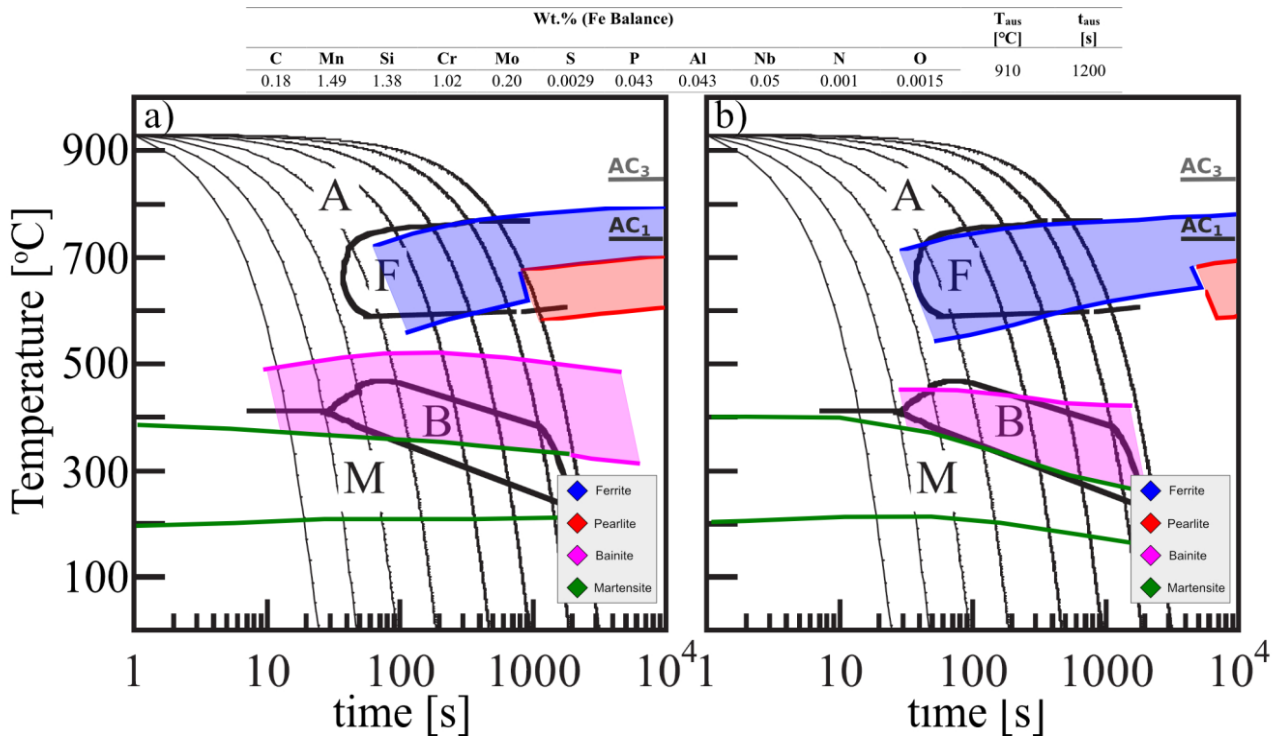

**Figure 8:** Predicted CCT diagram overlayed to experimental CCT [39] a) with no physics-informed integration b) with physics-informed integration.

**Impact of Physics-informed integration**

The effect of physics-informed integration on the accuracy of the predicted CCT diagrams is exemplified in Figure 8. The most significant improvements are observed in the bainitic transformation. It can be seen in the figure, for the slowest cooling rate, that in the model without PI integration (Figure 8a), the bainite start regressor lacks information about the previous formation of ferrite, leading to overestimated Bs temperatures compared to experimental observations. From a metallurgical standpoint, if no austenite had transformed in the range of 500°C, the supersaturation in the matrix would promote the formation of bainite at higher temperatures. In practice, the partial transformation of austenite into ferrite at higher temperatures, reduces matrix supersaturation and delays bainitic transformation. In this context, when the regressor for bainitic transformation range (ΔB) is coupled with bainite start temperature (Bs), its prediction gains accuracy. If bainitic transformation starts at lower temperatures, and less austenite is remaining for transformation, the PI-predicted bainitic range shortens. Such interaction between bainitic transformation and prior ferritic transformation is specifically visible at the slowest cooling rate, where the PI-integrated model (Figure 8b) correctly suppresses bainite formation once ferritic transformation has occurred. Thus, the integration of the phase fraction NN output with the bainitic transformation classifier, allows to refine the slow-end critical cooling rates for bainite.

Although the improvement is less pronounced, the introduction of synthetic data between discrete experimental cooling rates also enhances the phase formation classifiers' performance. Specifically, the accuracy of the predicted critical cooling rates for the fast cooling rates improve for ferrite, pearlite and bainite when PI integration is implemented. In the example shown, this refinement prevents the incorrect prediction of pearlite formation at the slowest cooling rate, an event not observed experimentally. Finally, as a consistency check, the predicted ferrite and pearlite start temperatures (Fs and Ps) remain below Ac3 and Ac1, respectively across all cooling rates. The modeling scheme thus produce transformation start temperatures consistent with physical metallurgy expectations, with Fs and Ps approaching their equilibrium values as the cooling rate decreases.

**Integrated CCT predictions compared to quenching dilatometry testing**

To further validate the CCT modeling, model predictions were also compared to data from new experiments. Through the new experiments we could ensure that all experimental conditions were well known, which adds additional reliability to the validation. The experimental validation consisted in a 2x2 factorial experiment evaluating the effects of C and Cr contents, while the rest of the elemental compositions were kept constant. The low values for C and Cr were 0.15 and 1 wt.%, and the high values were 1 and 4 wt.% respectively (see Table 2). This resulted in four alloys for quenching dilatometry tests, covering a relatively wide range of C and Cr contents that include hypo and hyper-eutectoid compositions.

A quenching and deformation dilatometer (Linseis, DIL L78RITA) and a high-temperature dilatometer (Netzsch DIL 402 SUPREME Expedis) were used in combination to detect the phase transformation temperatures from a wide range of cooling conditions including 9 cooling rates spanning from 0.083 to 150 °C/s. The steel specimens were machined to a geometry of 3×3×10 mm for the dilatometry measurements. The steels were heated from room temperature up to an austenitization temperature of 1050 °C, at a heating rate of 0.8 °C/s. This heating rate is sufficiently slow to consider the measured $Ac_1$, $Ac_{cm}$ and $Ac_3$ temperatures to be close to the equilibrium states. After reaching the austenitization temperature, the samples were immediately cooled to room temperature without holding. An example result from the dilatometry tests is presented in Figure 9. The onset and offset temperatures for various phase transformations were identified in the dilatometry curve by using the two-tangent line method.

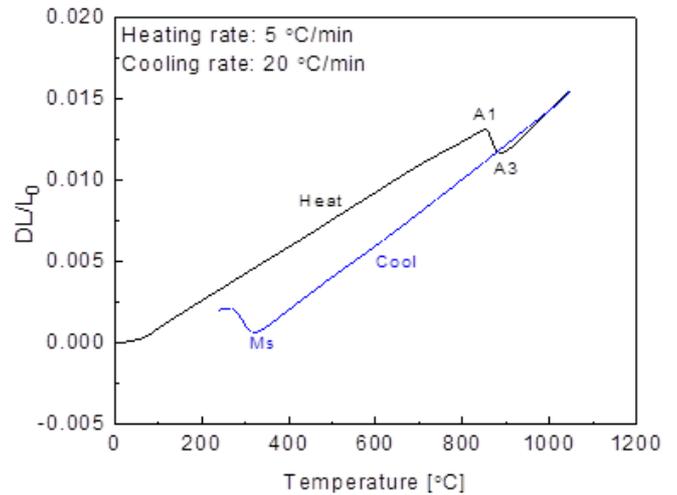

**Figure 9:** Example dilatometry curve with Ac1, Ac3 and Ms characteristic temperatures marked.

**Table 2:** Chemical composition of the investigated alloys (wt.%).

| Sample ID | C wt.% | Cr | Si | Mn | S | Al | Cu | Ni | Fe |
|---|---|---|---|---|---|---|---|---|---|
| A (0.15C 1Cr) | 0.14 | 0.98 | 0.02 | 0.07 | 0.06 | 0.01 | 0.01 | 0.017 | Bal. |
| B (0.15C 4Cr) | 0.16 | 4.05 | 0.02 | 0.08 | 0.05 | 0.02 | 0.009 | 0.015 | Bal. |
| C (1C 1Cr) | 0.95 | 1.06 | 0.02 | 0.07 | 0.09 | 0.03 | 0.01 | 0.017 | Bal. |
| D (1C 4Cr) | 0.88 | 4.12 | 0.02 | 0.08 | 0.05 | 0.02 | 0.005 | 0.012 | Bal. |

Figure 10 shows the experimentally measured $Ac_1$ and $Ac_3$ in relation to the equilibrium (calculated using Thermo-Calc Software TCFE10 [40]) and ML predicted critical transformation temperatures for all samples, see Table 3 for chemical compositions. Transformation temperatures determined under equilibrium conditions set a lower bound for such temperatures. Under equilibrium conditions the heating rate tends to zero, leading to no superheating needed for the transformation to occur. Under experimental conditions where the heating rate is aimed to be slow (but not zero) all the measured temperatures are equal to or higher than the equilibrium temperatures. For all samples, except for sample B, the measured $Ac_1$ and $Ac_3$ are higher than the one determined for equilibrium. In the case of sample B, the measured temperatures match the temperatures determined by equilibrium. The modelled transition temperatures lie very closely to equilibrium values (within ±1 °C), showing differences smaller than the experimental equipment's resolution. These conditions give modeled temperatures that are systematically lower than the experimentally measured temperatures. However, the mean absolute errors (MAE) between the experimental and modelled $Ac_1$ and $Ac_3$ are 15.3 and 14.2 °C, respectively. Considering experimental resolution and the uncertainties associated with the two-tangent line method for determining critical temperatures, a MAE below 20 °C indicates that the model shows good predictive capability. Such a level of accuracy is considered suitable for industrial applications, where temperature control is generally less precise than in laboratory-scale experiments.

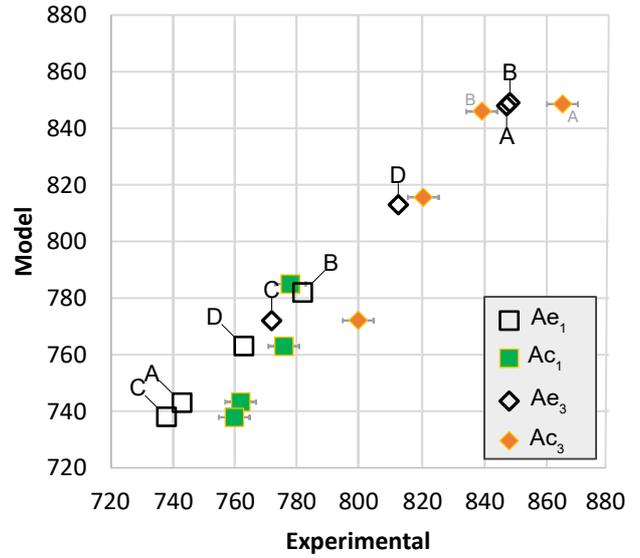

**Figure 10:** Scatter plot showing the correlation between the experimentally measured and modeled AC1 and AC3 temperatures for samples A, B, C and D. The predicted Ae1 and Ae3 under equilibrium conditions are also represented (predicted value on both x- and y-axis) for each alloy.

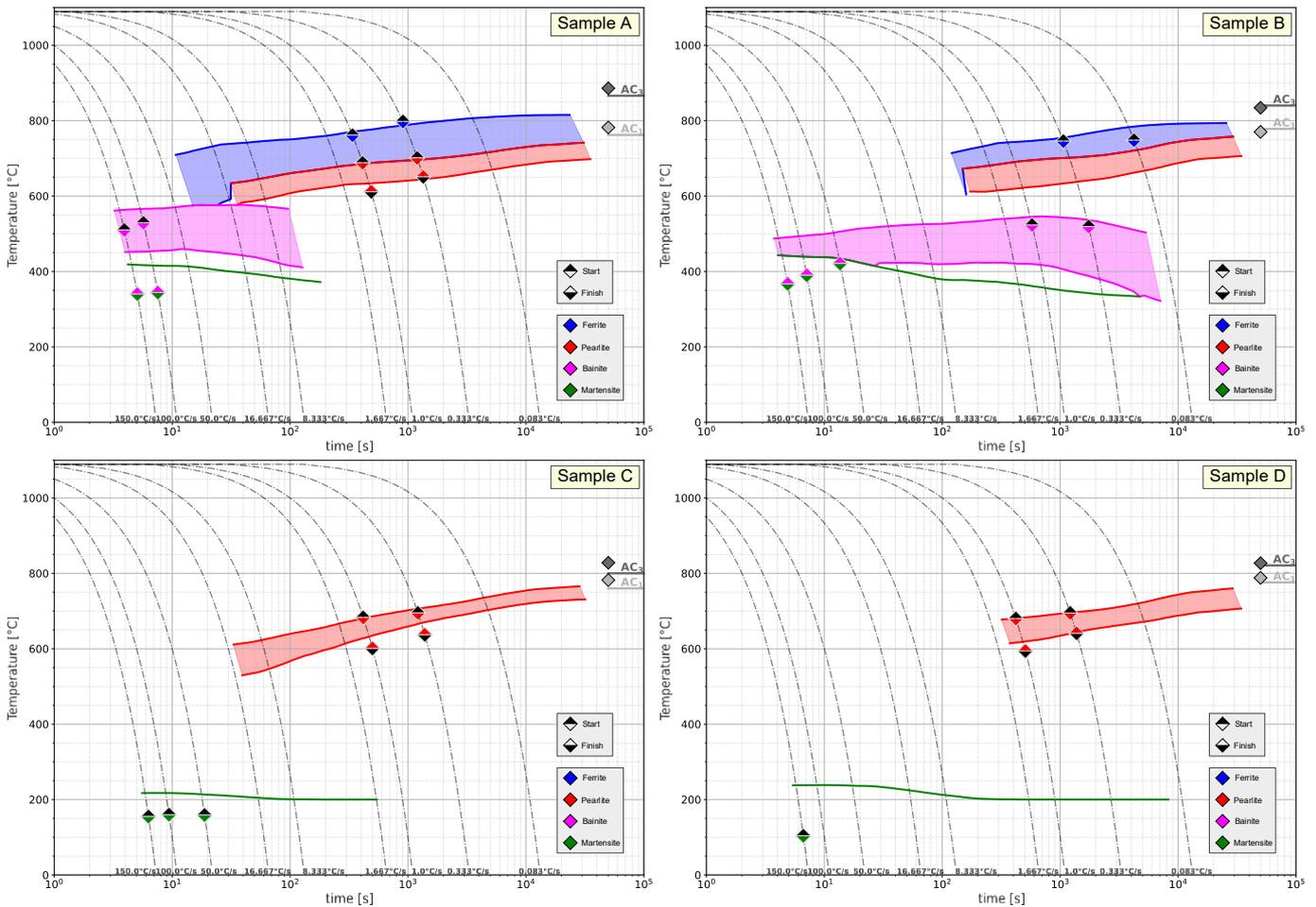

**Figure 11:** Calculated CCT diagrams for sample A, B, C and D with experimentally measured transition temperatures for ferrite, pearlite, bainite and martensite.

The good performance of the model in predicting $Ac_1$ and $Ac_3$ temperatures (Figure 10) served as a foundation for the integrating of characteristic temperature models to improve the predictive capability of the overall CCT model. The physics-informed ML for predicting the onset of ferrite and pearlite was conducted by tuning the error function to match the predicted $Ac_3$ and $Ac_1$ temperatures at cooling rates that approach zero.

Figure 11 shows the predicted CCT diagrams for the samples in Table 2, compared with experimentally measured cooling curves. The performance of the CCT modeling tool can be assessed by evaluating the predictability of the classification and regression models that the CCT model framework entail. For all the studied samples most of the phase transitions that are measured in the experiments are accurately predicted in the modeled CCT. The only exceptions are found in the measurement of pearlite at slow cooling rates for sample A and the measurement of martensite at high cooling rates for sample B.

In both cases, the transformation onset was challenging to measure separately from the ongoing transformation. In the case of sample A, it is reasonable to assume that the formation of martensite starts close to, or at the points, labeled as bainitic transformation completion. Regarding sample B, the ongoing ferritic transformation makes the precise determination of the limit between the ferritic and pearlitic transformation difficult. Despite these difficult cases, it is relevant to highlight the performance of the model in the prediction of samples C and D, with high C content. The promotion of pearlite formation instead of the formation of ferrite can be easily predicted by simply considering the hyper-eutectoid region in the Fe-C phase diagram. In such region, and for higher carbon contents, the formation of pro-eutectoid cementite would not be predicted by the current modelling framework. A natural extension of the current work's scope would be the training of classification and regression models for cementite using data from high carbon steels diagrams. The modeled CCT diagrams for samples C and D accurately predict the formation of pearlite and the absence of ferrite. Additionally, experimental measurements did not detect the formation of pro-eutectoid cementite in any of the measured samples. Unlike pearlite or ferrite, the formation of metastable microconstituents such as bainite or martensite can't be assessed using equilibrium phase diagrams; and suitable CCT diagrams become crucial to study their stability regions. The presented model is also accurate at predicting for both samples the absence of experimentally measured bainite, as well as the formation of martensite.

The predictions of the classification models can also be analyzed to study the effect that C or Cr variation have on the stability of the forming microconstituents. A commonly accepted trend analysis is that more alloying content added to the base iron will displace the transformation curves towards slower cooling rates. Indeed, the sample with a slower critical cooling rate is sample D (1.2 °C/s), which possess the highest alloying among the studied samples (Figure 11.d). When comparing sample A and B it is clear how increasing the Cr content displaces the ferrite and pearlite curves towards longer times. The same is observed in samples C and D, where the high C content has inhibited the formation of ferrite, and the increase of Cr content displaces the pearlitic transformation onset from a cooling rate of 13.5 to 1.2 °C/s. These observations are in good agreement with the standard categorization of C and Cr as austenite and ferrite stabilizing elements, respectively [41].

The reasonable accuracy achieved by the classification models on the phase formations in the studied alloys enables the analysis of the accuracy of the regression models to predict the transformation temperatures upon cooling. Table 3 shows the MAE between predicted and measured transformation start temperatures for ferrite, pearlite, bainite and martensite for all samples. All regression models except for the one predicting martensitic transformation show a reasonable accuracy lower than or close to ±20°C. Specifically, the models trained for the prediction of ferrite and pearlite show very high accuracies for the studied alloys. On the other hand, the prediction of bainite and martensite start temperatures show a lower accuracy, with a systematic prediction of a higher transformation start temperature than the ones measured experimentally. For bainite, the model still presents reasonable predictions (±30°C). However, for martensite the error is larger and thus further iterations on the modelling strategies, and the dataset used to train the model would be of interest in future work. The higher measured error is found when modeling sample D, containing high carbon and chromium contents. These results already suggest a direction for future work aimed at improving accuracy, as in such alloys, modeling additional microconstituents beyond ferrite, pearlite, bainite or martensite, such as carbides, may be important to increase the model's predictive capability. Hence, extending the used dataset to include steels grades with higher alloy contents and accounting for the presence and type of carbides at austenitization by considering i.e. austenite compositions (rather than nominal compositions) could significantly improve the model's accuracy.

In addition to the comparison of the measured transformation temperatures with their corresponding predicted temperatures, the use of computational tools allows for the exploration of broader compositional landscapes without requiring extensive experimental efforts. In this context, not only individual steels can be evaluated but also the impact of varying the concentration of individual elements on the phase transformation temperatures can be assessed.

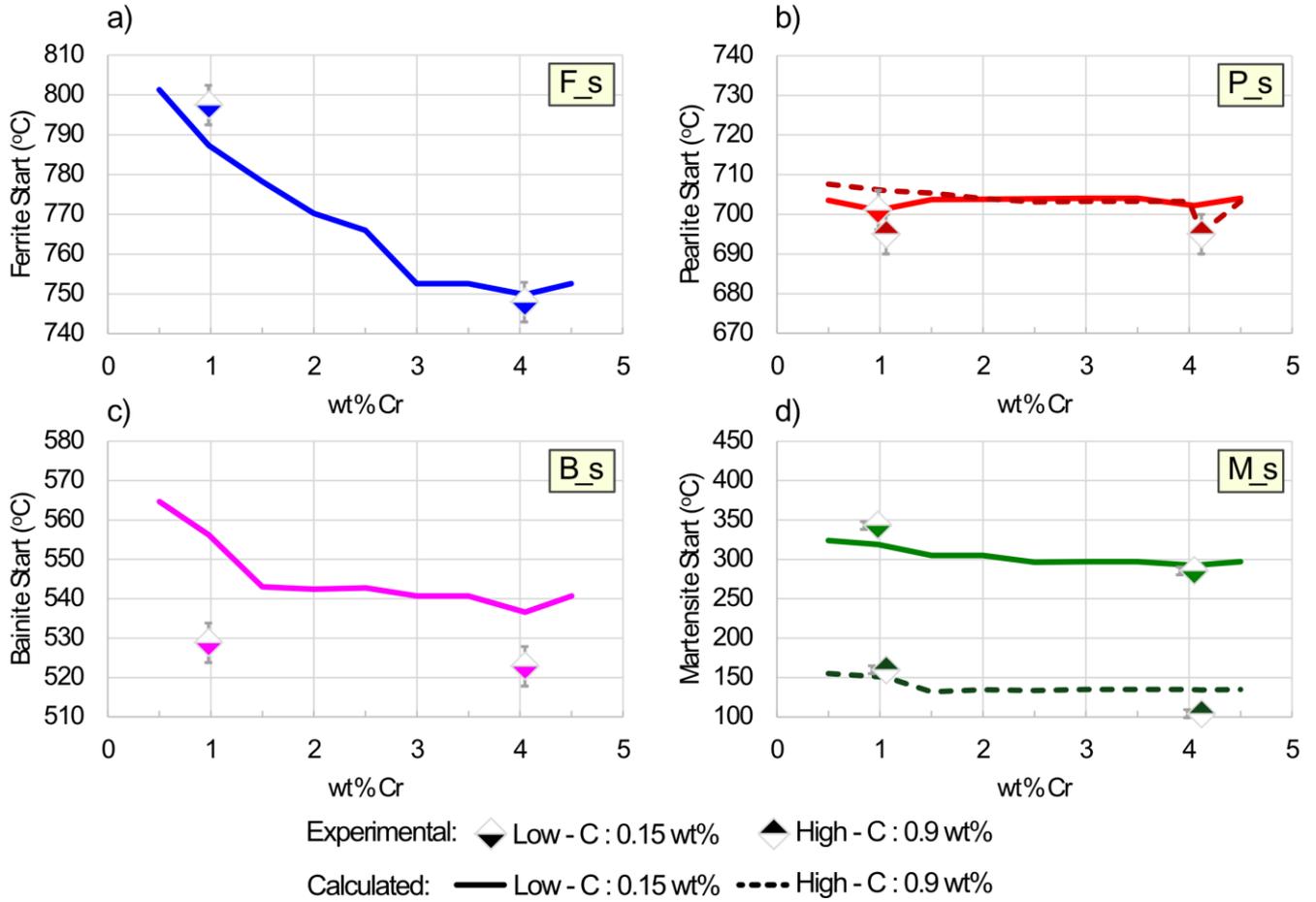

**Figure 12:** Effect of Cr content on phase transformation start temperature for a) ferrite, b) pearlite, c) bainite and d) martensite for low and high C containing grades. Fixed cooling rates for Ferrite, Pearlite, Bainite and Martensite start temperatures are 0.33, 0.33, 50 and 50 oC/s respectively.

Following the 2x2 experimental design, the effect of varying Cr content on transformation temperatures is assessed for the two carbon levels at a fixed cooling rate for each constituent. The experimental compositions of samples A, B, C and D are explicitly used in the simulations, while for the rest of the composition iterations, when calculating low-C grades the composition of each element is taken as an average value between sample A and B (except for Cr that is changed in steps of 0.5 wt.%). The same approach is applied to the high-C grades but using the average elemental compositions of samples C and D.

Figure 12 presents predicted and experimental transformation temperatures resulting from the described compositional variations. The model accurately captures the effect of increasing Cr content on lowering the ferrite start temperature. Regarding pearlitic transformation, experiments show that Cr has a small impact on the pearlite start temperature, a trend also reflected in the simulations for both carbon levels. Regarding bainite and martensite transformations, while individual predictions may be biased (as previously discussed), the model successfully reproduces the general trends established by the experimental measurements. Specifically for martensite, the model slightly underestimates the decrease in transformation onset with increasing Cr content. This suggests that implementing the previously discussed Ms model refinements could enhance its predictive capability also for martensitic transformation trends with compositional variations.

**Table 3:** Mean absolute error for the steels listed in Table 2.

| MAE | Low-Cr: 1 wt.% Cr | | High-Cr: 4 wt.% Cr | |
|---|---|---|---|---|
| Low-C: | SAMPLE A: | | Sample B: | |
| 0.15 wt.% C | $\Delta F_s$ | 8.4 | $\Delta F_s$ | 18.1 |
| | $\Delta P_s$ | 5.0 | $\Delta P_s$ | - |
| | $\Delta B_s$ | 43.8 | $\Delta B_s$ | 19.8 |
| | $\Delta M_s$ | 75.5 | $\Delta M_s$ | 46.6 |
| High-C: | sample C: | | Sample D: | |
| 0.9 wt.% C | $\Delta F_s$ | | $\Delta F_s$ | |
| | $\Delta P_s$ | 6.4 | $\Delta P_s$ | 0.5 |
| | $\Delta B_s$ | | $\Delta B_s$ | |
| | $\Delta M_s$ | 58.0 | $\Delta M_s$ | 133.7 |

The proposed physics-informed ML modeling framework for predicting CCT diagrams offers a key advantage over traditional physically based simulations and especially over experimental methods. In experimental CCT measurements, the discrete nature of the cooling rates that can be tested introduces uncertainty in determining the precise starting times of phase transformations. Accurately identifying phase formation onset requires a significant increase in the number of cooling experiments, leading to higher time and cost requirements. Furthermore, the detection of phase transformation limits is constrained by the resolution of the experimental equipment, which typically struggles to identify phase fractions below approximately 1% by volume. In contrast, once the predictive accuracy of the ML-based models is validated, they can provide continuous phase transformation limits without additional experimental costs or time.

## Summary and Outlook

In this work we have introduced a physics-informed machine learning (ML) framework for modeling continuous cooling transformation (CCT) behavior in steels, addressing a key scientific gap in current ML applications for steel: the lack of physically grounded strategies and the reliance on small, fragmented datasets. While ML has proven effective in materials discovery and process optimization, its use in general-purpose modeling for complex industrial materials like steel remains limited. Our approach combines metallurgical domain knowledge with scalable ML techniques to improve predictive accuracy, physical consistency, and computational efficiency.

The CCT model developed here is trained on a dataset of unprecedented scale, 4100 CCT diagrams, and validated against both literature data and new experimental dilatometry measurements. It demonstrates high computational efficiency, generating full CCT diagrams in approximately 5 seconds. The model achieves strong generalizability across low-alloy steels, with phase formation classification F1 scores exceeding 88% for all phases. For phase transformation temperature regression, mean absolute errors (MAE) are below 20 °C for all phases except bainite, which shows a slightly higher MAE of 27 °C.

The model architecture incorporates physics-informed strategies such as transformation sequence constraints, thermodynamic coherence, and feature engineering based on metallurgical kinetics. These elements ensure realistic predictions and improve robustness across a wide compositional range. While key experiments remain essential, the framework enables virtual testing that can reduce the need for large experimental campaigns and accelerate alloy and process development.

Looking ahead, the framework provides a strong foundation for building a universal digital twin for heat treatment and related steel processing applications. Future developments will focus on expanding the architecture with additional generic and customized ML models, enabling broader applicability across steel grades and consideration of complete processing routes. Integration with complementary simulation tools, such as finite element method (FEM) models for thermal conditions and CALPHAD-based equilibrium calculations for phase balance and composition prior to cooling, could already be applied to simulate, for example, industrial heat treatment of highly alloyed steels where secondary phases may exist prior to cooling. It should be noted that this comprehensive capability can also be implemented directly in the physics-informed ML framework for heat treatment going forward.

Feature engineering will continue to play a central role in improving ML model accuracy. For example, incorporating prior austenite grain size directly, rather than relying solely on austenitization temperature and time, could strengthen correlations with transformation behavior. The framework also offers potential for handling subtle compositional variations between production batches, which is critical for robust industrial heat treatment. When embedded within digital twin environments, the model can be adapted to site-specific conditions, enabling expert-level predictions tailored to local manufacturing workflows.

As the steel industry faces increasing pressure to innovate and decarbonize, physics-informed ML frameworks like this one offer a scientifically grounded and computationally efficient path forward. By integrating physical insight with scalable data-driven tools, this approach supports accelerated materials design, process optimization, and implementation across the steel value chain.

## Acknowledgements

The authors would like to acknowledge the contributions to this work by Felix Rios, Ismael Serrano Garcia, Vishnu Sharma and Moshiour Rahaman. The financial contributions from Vinnova and the EIT Raw Materials project ENDUREIT (No. 18317) are gratefully acknowledged.

## Declaration of Generative AI in the writing process

At the final stage of the writing process, the authors used ChatGPT to improve readability and language of the manuscript. After using this tool/service, the authors reviewed and edited the content as needed and take full responsibility for the content of the article.

## References

1. J. Allison, D. Backman, and L. Christodoulou: JOM, 2006, vol. 58, pp. 25–27. https://doi.org/10.1007/s11837-006-0223-5.
2. G.B. Olson: Acta Mater., 2013, vol. 61, pp. 771–781. https://doi.org/10.1016/j.actamat.2012.10.045.
3. L. Kaufman and J. Ågren: Scr. Mater., 2014, vol. 70, pp. 3–6. https://doi.org/10.1016/j.scriptamat.2012.12.003.
4. L. Vitos, J.-O. Nilsson, and B. Johansson: Acta Mater., 2006, vol. 54, pp. 3821–3826. https://doi.org/10.1016/j.actamat.2006.04.013.
5. H.K.D.H. Bhadeshia, D.J.C. MacKay, and L.E. Svensson: Mater. Sci. Technol., 1995, vol. 11, pp. 1046–1051. https://doi.org/10.1179/mst.1995.11.10.1046.
6. H.K.D.H. Bhadeshia: ISIJ Int., 1999, vol. 39, pp. 966–979. https://doi.org/10.2355/isijinternational.39.966.
7. W.G. Vermeulen, P.J. van der Wolk, A.P. de Weijer, and S. van der Zwaag: J. Mater. Eng. Perform., 1996, vol. 5, pp. 57–63. https://doi.org/10.1007/BF02647270.
8. G.L.W. Hart, T. Mueller, C. Toher, and S. Curtarolo: Nat. Rev. Mater., 2021, vol. 6, pp. 730–755. https://doi.org/10.1038/s41578-021-00340-w.
9. G.E. Karniadakis, I.G. Kevrekidis, P. Perdikaris, S. Wang, and L. Yang: Nat. Rev. Phys., 2021, vol. 3, pp. 422–440. https://doi.org/10.1038/s42254-021-00314-5.
10. C. Wang, K. Zhu, P. Hedström, Y. Li, and W. Xu: J. Mater. Sci. Technol., 2022, vol. 128, pp. 31–43. https://doi.org/10.1016/j.jmst.2022.04.014.
11. Y. Cao, C. Zhang, S. Tang, S. Wu, X. Zhou, G. Cao, D. Luo, H. Wang, P. Hedström, and Z. Liu: Mater. Des., 2025, vol. 258, 114642. https://doi.org/10.1016/j.matdes.2025.114642.
12. J.A. Lee, R.B. Figueiredo, H. Park, J.H. Kim, and H.S. Kim: Acta Mater., 2024, vol. 275, 120046. https://doi.org/10.1016/j.actamat.2024.120046.
13. B. Zhang, B. Wang, W. Xue, A. Ullah, T. Zhang, and H. Wang: J. Mater. Sci., 2023, vol. 58, pp. 4795–4808. https://doi.org/10.1007/s10853-023-08322-9.

14. S. Minamoto, S. Tsukamoto, T. Kasuya, M. Watanabe, and M. Demura: Sci. Technol. Adv. Mater.: Methods, 2022, vol. 2(1), pp. 402–415. https://doi.org/10.1080/27660400.2022.2123262.
15. S. Ganguly and S. Manna: Mater. Manuf. Process., 2023, vol. 38(16), pp. 2018–2033. https://doi.org/10.1080/10426914.2023.2190388.
16. W. Trzaska: Arch. Metall. Mater., 2018, vol. 63(4), pp. 2009–2015. https://doi.org/10.24425/amm.2018.125137.
17. J. Miettinen, S. Koskenniska, M. Somani, S. Louhenkilpi, A. Pohjonen, J. Larkiola, and J. Kömi: Metall. Mater. Trans. B, 2019, vol. 50, pp. 2853–2866. https://doi.org/10.1007/s11663-019-01698-7.
18. H. Martin, P. Amoako-Yirenkyi, A. Pohjonen, N.K. Frempong, J. Kömi, and M. Somani: Metall. Mater. Trans. B, 2021, vol. 52(2), pp. 223–235. https://doi.org/10.1007/s11663-020-01991-w.
19. Y. Cao, G. Cao, C. Cui, X. Li, S. Wu, and Z. Liu: Metall. Mater. Trans. A, 2023, vol. 54A(12), pp. 4891–4904. https://doi.org/10.1007/s11661-023-07210-w
20. SteelData: steeldata.info, 2019.
21. S. Gojare, R. Joshi, and D. Gaigaware: Procedia Computer Science, 2015, vol. 50, pp.341-346.
22. M. Abadi et al.: https://www.tensorflow.org/, 2015.
23. F. Pedregosa, G. Varoquaux, A. Gramfort, V. Michel, B. Thirion, O. Grisel, M. Blondel, P. Prettenhofer, R. Weiss, V. Dubourg, J. Vanderplas, A. Passos, D. Cournapeau, M. Brucher, M. Perrot, and E. Duchesnay: J. Mach. Learn. Res., 2011, vol. 12, pp. 2825–2830.
24. T. Chen and C. Guestrin: Proc. 22nd ACM SIGKDD Int. Conf. Knowl. Discov. Data Min., San Francisco, CA, USA, 2016. https://doi.org/10.1145/2939672.2939785.
25. T. Hastie and R. Tibshirani: Adv. Neural Inf. Process. Syst., 1995, vol. 8.
26. L. Devroye, L. Györfi, and G. Lugosi: Springer Sci. Bus. Media, 2013, vol. 31.
27. S. Kunapuli: Manning, 2023, vol. 13. ISBN: 9781617297137.
28. T. Hastie, R. Tibshirani, and J.H. Friedman: Springer, 2009, vol. 2. https://doi.org/10.1007/978-0-387-84858-7
29. Z. Babasafari, A.V. Pan, F. Pahlevani, R. Hossain, V. Sahajwalla, M. Du Toit, and R. Dippenaar: J. Mater. Res. Technol., 2020, vol. 9(6), pp. 15286–15297. https://doi.org/10.1016/j.jmrt.2020.10.071.
30. A. Di Schino, M. Gaggiotti, and C. Testani: Metals, 2020, vol. 10(6), 808. https://doi.org/10.3390/met10060808.
31. Gramlich, C. van der Linde, M. Ackermann, and W. Bleck: Results Mater., 2020, vol. 8, 100147. https://doi.org/10.1016/j.rinma.2020.100147
32. S.K. Md Hasan, M. Ghosh, D. Chakrabarti, and S.B. Singh: Mater. Sci. Eng. A, 2020, vol. 771, 138590. https://doi.org/10.1016/j.msea.2019.138590.
33. R. Kawulok, I. Schindler, P. Kawulok, S. Rusz, P. Opěla, Z. Solowski, and K.M. Čmiel: Metalurgija, 2015, vol. 54(3), pp. 473–476.
34. R. Kawulok, I. Schindler, J. Sojka, P. Kawulok, P. Opěla, L. Pindor, E. Grycz, S. Rusz, and V. Ševčák: Crystals, 2020, vol. 10(4), art. 326. https://doi.org/10.3390/cryst10040326.
35. R. Neugebauer, A. Rautenstrauch, and E.M. Garcia: Adv. Mater. Res., 2011, vol. 337, pp. 358–362. https://doi.org/10.4028/www.scientific.net/AMR.337.358.
36. Y. Pei, Z. Gao, Y. Liu, S.Q. Zhao, C.Y. Xu, L.Y. Ren, and X.L. Li: Adv. Mater. Res., 2014, vol. 1035, pp. 27–35. https://doi.org/10.4028/www.scientific.net/AMR.1035.27.
37. E.J. Seo, L. Cho, and B.C. De Cooman: Metall. Mater. Trans. A, 2014, vol. 45, pp. 4022–4037. https://doi.org/10.1007/s11661-014-2657-7.
38. J.M. Tartaglia, A.N. Kuelz, and V.H. Thelander: J. Mater. Eng. Perform., 2018, vol. 27(12), pp. 6349–6364. https://doi.org/10.1007/s11665-018-3683-1.
39. J. Kobayashi, D. Ina, N. Yoshikawa, and K. Sugimoto: ISIJ Int., 2012, vol. 52(10), pp. 1894–1901. https://doi.org/10.2355/isijinternational.52.1894.
40. Thermo-Calc Software: TCFE Steels/Fe-alloys Database version 10, https://thermocalc.com/products/databases/steel-and-fe-alloys
41. J.L. Dosset: ASM Int., 2020. ISBN: 978-1-62708-325-6.

**Table S.1:** Alloy compositions and mean absolute error for transformation temperatures for complementary literature benchmarking.

| | Wt.% (Fe Balance) | | | | | | | | | | | | | $T_{aus}$ [°C] | $t_{aus}$ [s] | Mean Abs. Error [°C] | | | | |
|---|---|---|---|---|---|---|---|---|---|---|---|---|---|---|---|---|---|---|---|---|
| ID | C | Mn | Si | Cr | Mo | S | P | Ni | Al | Nb | N | Ti | B | Other | | | F_s | P_s | B_s | M_s | Ref |
| 1 | 0.069 | 0.86 | 0.24 | 1.46 | 1.29 | | 0.02 | | | | | | | | 1000 | 1200 | 17.1 | 8.1 | 9.7 | 20.5 | (34) |
| 2 | 0.07 | 0.33 | 0.24 | 2.69 | 1.29 | | 0.01 | | | | | | | | 1000 | 1200 | 19.1 | | 5.8 | | (34) |
| 3 | 0.071 | 0.37 | 0.26 | 1.53 | 0.63 | | 0.009 | | | | | | | | 1000 | 1200 | 7.4 | 18.5 | 39.4 | | (34) |
| 4 | 0.072 | 0.91 | 0.28 | 2.78 | 0.63 | | 0.021 | | | | | | | | 1000 | 1200 | 10.8 | | 7.3 | 15.6 | (34) |
| 5 | 0.094 | 2.07 | 0.276 | 0.75 | 0.245 | 0.015 | 0.036 | | | | | 0.03 | 0.005 | | 900 | 600 | 8.5 | 11.6 | 14.3 | 9.9 | (31) |
| 6 | 0.11 | 9.94 | 0.15 | 0.11 | 0.02 | | | 0.1 | 0.021 | 0.01 | | 0.003 | 0.0005 | | 747 | 600 | | | | 79.7 | (27) |
| 7 | 0.15 | 1.5 | 0.5 | | | 0.002 | 0.015 | | | | | | | | 900 | 600 | 16.9 | 24.3 | 34.7 | 26.4 | (31) |
| 8 | 0.15 | 1.58 | 0.55 | 0.022 | 0.002 | 0.007 | 0.03 | 0.01 | | 0.033 | | | | | 980 | 600 | 8.5 | 52.6 | 40.2 | 43.3 | (32) |
| 9 | 0.15 | 4.02 | 0.49 | 0.12 | 0.2 | | | 0.1 | 0.027 | 0.035 | | 0.003 | 0.0005 | | 856 | 600 | | | 74.6 | 6.2 | (27) |
| 10 | 0.154 | 1.1 | 0.022 | 0.05 | | 0.0046 | | | 0.043 | | 0.005 | 0.002 | | Ca:0.003 | | 1 | 6.0 | 4.5 | 15.5 | 6.8 | (13) |
| 11 | 0.16 | 0.86 | 0.27 | 1.46 | 0.62 | | 0.009 | | | | | | | | 1000 | 1200 | 10.9 | | 5.7 | 30.9 | (34) |
| 12 | 0.16 | 4 | 0.52 | 0.11 | 0.21 | | | 0.1 | 0.51 | 0.037 | | 0.003 | 0.003 | | 918 | 600 | 28.9 | | 52.8 | 19.9 | (27) |
| 13 | 0.17 | 0.33 | 0.26 | 2.73 | 0.62 | | 0.022 | | | | | | | | 1000 | 1200 | 53.6 | 23.3 | 3.8 | 51.0 | (34) |
| 14 | 0.17 | 0.36 | 0.26 | 1.46 | 1.25 | | 0.021 | | | | | | | | 1000 | 1200 | 10.3 | 22.1 | 9.0 | 22.6 | (34) |
| 15 | 0.17 | 0.85 | 0.24 | 2.78 | 1.27 | | 0.009 | | | | | | | | 1000 | 1200 | 25.2 | | 13.1 | 17.5 | (34) |
| 16 | 0.17 | 3.96 | 0.5 | 0.12 | 0.02 | | | 0.1 | 0.027 | 0.032 | | 0.003 | 0.0005 | | 852 | 600 | 21.6 | | 61.9 | 4.2 | (27) |
| 17 | 0.17 | 3.99 | 0.5 | 0.11 | 0.02 | | | 0.1 | 0.025 | 0.033 | | 0.02 | 0.0057 | | 860 | 600 | | | 65.7 | 2.6 | (27) |
| 18 | 0.18 | 1.36 | 1.2 | 1.4 | 0.49 | 0.011 | 0.027 | | | | | | | | 1000 | | | 3.0 | 14.8 | 47.3 | (28) |
| 19 | 0.18 | 2.2 | 1 | 0.6 | 0.4 | 0.01 | 0.025 | | | | | 0.18 | | | 900 | 600 | 82.9 | 23.7 | 24.7 | 74.0 | (31) |
| 20 | 0.19 | 4.02 | 0.5 | 0.11 | 0.02 | | | 0.09 | 0.031 | 0.035 | | 0.02 | 0.0016 | | 842 | 600 | | | 57.2 | 4.7 | (27) |
| 21 | 0.21 | 6.55 | 0.16 | 0.1 | 0.02 | | | 0.11 | 0.025 | 0.01 | | 0.003 | 0.0005 | | 786 | 600 | | | | 7.2 | (27) |
| 22 | 0.23 | 1.52 | 1.48 | 1.2 | 0.35 | 0.01 | 0.024 | 0.85 | | | | | | | 1000 | | | 13.7 | 3.9 | 7.2 | (28) |
| 23 | 0.24 | 1.14 | 0.27 | 0.17 | | | | | | | 0.001 | 0.036 | 0.003 | | 900 | 240 | 20.2 | | 21.6 | 11.0 | (33) |
| 24 | 0.251 | 1.3 | 0.238 | 0.113 | 0.006 | 0.016 | 0.03 | | | | | 0.037 | 0.005 | | 900 | 600 | 10.5 | 7.0 | 24.0 | 9.8 | (31) |
| 25 | 0.27 | 1.5 | 1.61 | 0.001 | | | | | | | 0.0026 | 0.027 | 0.0025 | | 900 | 240 | 9.1 | | 10.5 | 2.3 | (33) |
| 26 | 0.28 | 1.46 | 1.58 | 0.97 | | | | | | | 0.0025 | 0.023 | 0.0023 | | 900 | 240 | 18.4 | | 94.4 | 21.1 | (33) |
| 27 | 0.31 | 9.85 | 0.15 | 0.11 | 0.01 | | | 0.1 | 0.02 | 0.01 | | 0.003 | 0.0005 | | 747 | 600 | | | | 28.1 | (27) |
| 28 | 0.32 | 0.75 | 0.1 | 1.11 | | 0.008 | 0.01 | | | | | | 0.003 | Cu: 0.02 | 850 | 120 | 14.0 | 7.0 | 9.9 | 3.2 | (29) |
| 29 | 0.32 | 3 | 0.16 | 0.1 | 0.02 | | | 0.1 | 0.026 | 0.005 | | 0.003 | 0.0005 | | 826 | 600 | | | 28.9 | 4.9 | (27) |
| 30 | 0.42 | 0.65 | 0.5 | 7 | 0.7 | | | 0.2 | | | | | | V:0.1 | 1200 | 3600 | | 4.2 | 12.9 | 11.1 | (26) |
| 31 | 0.42 | 0.65 | 0.5 | 7 | 0.7 | | | 0.2 | | | | | | V:0.1 | 980 | 3600 | | 4.1 | 22.6 | 19.4 | (26) |
| 32 | 0.42 | 0.68 | 0.26 | 1.05 | 0.21 | | | 0.17 | 0.028 | 0.001 | | 0.001 | 0.0005 | | 842 | 600 | 7.3 | | 12.5 | 4.9 | (27) |
| 33 | 0.994 | 0.38 | 0.324 | 1.45 | | 0.01 | 0.011 | | | | | | | | 850 | 600 | | 19.6 | | 5.3 | (30) |
| 34 | 1 | 0.98 | 0.21 | 0.63 | 0.024 | | | 0.09 | | | | | | Cu: 0.18 | 1000 | 600 | | 2.8 | 40.5 | 10.7 | (25) |
| | | | | | | | | | | | | | | | | Average: | 19.4 | 14.7 | 27.7 | 19.7 | |